\documentclass[runningheads]{llncs}
\usepackage{graphicx}
\usepackage{amsmath,amssymb} % define this before the line numbering.
\usepackage{color}
\usepackage[width=122mm,left=12mm,paperwidth=146mm,height=193mm,top=12mm,paperheight=217mm]{geometry}
\usepackage{booktabs}
\usepackage{url}
\usepackage[utf8]{inputenc}
\begin{document}
% \linenumbers
\pagestyle{headings}
\mainmatter

\title{Deep Impression: Audiovisual Deep Residual Networks for Multimodal Apparent Personality Trait Recognition} % Replace with your title

\titlerunning{Deep Impression}

\authorrunning{Güçlütürk et al.}

\author{Yağmur Güçlütürk, Umut Güçlü, Marcel A. J. van Gerven, Rob van Lier}

%Please write out author names in full in the paper, i.e. full given and family names. 
%If any authors have names that can be parsed into FirstName LastName in multiple ways, please include the correct parsing, in a comment to the volume editors:
%\index{Lastnames, Firstnames}
%(Do not uncomment it, because you may introduce extra index items if you do that...)

\institute{Radboud University, Donders Institute for Brain, Cognition and Behaviour, Nijmegen, the Netherlands\\
	\email{ \{y.gucluturk, u.guclu, m.vangerven, r.vanlier\}@donders.ru.nl}
}

\maketitle

\begin{abstract}
Here, we develop an audiovisual deep residual network for multimodal apparent personality trait recognition. The network is trained end-to-end for predicting the Big Five personality traits of people from their videos. That is, the network does not require any feature engineering or visual analysis such as face detection, face landmark alignment or facial expression recognition. Recently, the network won the third place in the ChaLearn First Impressions Challenge with a test accuracy of 0.9109.
\end{abstract}\section{Introduction}

Appearances influence what people think about the personality of other people, even without having any interaction with them. These judgments can be made very quickly - already after 100 ms \cite{Willis2006}. Although some studies have shown that people are good at forming accurate first impressions about the personality traits of people after viewing their photographs or videos \cite{Borkenau1992,Naumann2009}, it has also been shown that simply relying on the appearance does not always result in correct first impression judgments \cite{Olivola2010}.

Several characteristics of people varying from clothing to facial expressions, contribute to the first impression judgments about personality \cite{TeijeiroMosquera2015}. For example, \cite{Todorov2014} has shown that the photographs of the same person taken with a different facial expression changes the judgments about the person's personality traits such as trustworthiness and extravertedness as well as other perceived characteristics such as attractiveness and intelligence.  Furthermore, people are better at guessing other's personality traits if they find them attractive after short encounters with them \cite{Lorenzo2010}. The same study also showed that people form more positive first impressions about more attractive people.

Studies of personality prediction generally either deal with correctly recognizing the actual personality traits of people, which can be measured as self- or acquaintance-reports or apparent personality traits, which are the impressions about the personality of an unfamiliar individual \cite{Vinciarelli2014a}. Below we review the recent work in apparent personality prediction.

% \section{Related Work}

Most of the previous work on apparent personality modeling and prediction have been in the domain of paralanguage, i.e. speech, text, prosody, other vocalizations and fillers  \cite{Vinciarelli2014a}. Conversations (both text and audio) \cite{Mairesse2007} and speech clips \cite{Polzehl2010,Mohammadi2015} were the materials that were most commonly analyzed. In this domain, INTERSPEECH 2012 Speaker Trait Challenge \cite{Schuller2015} enabled a systematic comparison of computational methods by providing a dataset comprising audio data and extracted features. The competition had three sub-challenges for predicting the Big Five personality traits, likability and pathology of speakers,  .

Recently, prediction of apparent personality traits from social media content has become a challenge that attracted much attention in the field. For example \cite{Cristani2013,Segalin2016} demonstrated that the images that the users "favorite" on Flickr enabled the prediction of both apparent and actual (self-assessed) personality traits of Flickr users. \cite{Vernon2014} looked at the influence of a large number of physical attributes (e.g. chin length, head size, posture) on people's impressions regarding approachability, youthful-attractiveness and dominance of them. They studied these influences based on people's impressions formed after looking at face photographs. They performed factor analysis to quantify the contribution of physical attributes and used these factors as inputs to a linear neural network to predict impressions. Their predictions were significantly correlated to the actual impression data.

Given that the exact facial expression \cite{Todorov2014} and the posture \cite{Vernon2014} of the person in a photograph influences the first impression judgments about that person, as well as the importance of paralinguistic information in impression formation \cite{Mairesse2007}, continuous audio-visual data seems to be a suitable medium to study first impressions. In a series of studies using YouTube video blogs (vlogs) \cite{Biel2011,Biel2012,Biel2013,TeijeiroMosquera2015}, researchers showed that this is indeed the case. Furthermore, \cite{Celiktutan2016} showed that audiovisual annotations along with audiovisual cues enabled the best prediction performance for their regression models compared to either using only either one of them.

At the same time, deep neural networks \cite{Schmidhuber2015,LeCun2015} in general and deep residual networks \cite{He2015} in particular have achieved state-of-the-art results in many computer vision tasks. For example, \cite{He2015} won the first places in the object detection task and the object localization task at the ImageNet Large Scale Visual Recognition Challenge 2015\footnote{\url{http://image-net.org/}.} with their seminal work that introduced deep residual networks. Furthermore, deep residual networks have been successfully used in a variety of other computer vision tasks ranging from style transfer \cite{Johnson2016} and image super-resolution \cite{Johnson2016} to semantic segmentation \cite{Dai2015} and face hallucination \cite{Gucluturk2016}.

Recently, \cite{Vinciarelli2014b} suggested that deep neural networks can be used for personality trait recognition because of the hierarchical organization of the personality traits \cite{Wright2014}. Following this line of reasoning as well as the recent success of deep residual networks, we develop an audiovisual deep residual network for multimodal personality trait recognition. The network is trained end-to-end for predicting the apparent Big Five personality traits of people from their videos. That is, the network does not require any feature engineering or visual analysis such as face detection, face landmark alignment or facial expression recognition.

\section{Methods}

\subsection{Architecture}

\begin{figure}[!ht]
  \centering
    \includegraphics[width=1\textwidth]{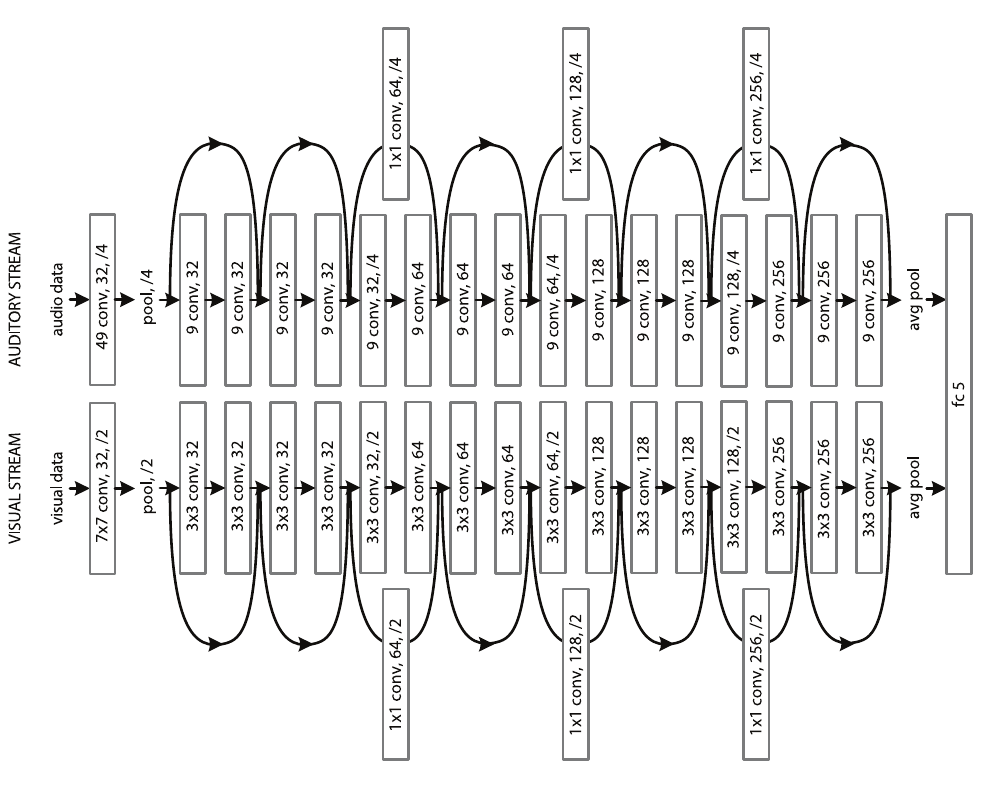}
  \caption{Illustration of the network architecture.}
  \label{figure_1}
\end{figure}

Figure \ref{figure_1} shows an illustration of the network architecture. The network comprises an auditory stream of a 17 layer deep residual network, a visual stream of another 17 layer deep residual network and an audiovisual stream of a fully-connected layer.

The auditory stream and the visual stream are similar to the first 17 layers of the 18 layer deep residual network in \cite{He2015}. That is, each stream comprises one convolutional layer and eight residual blocks of two convolutional layers. The convolutional layers are followed by batch normalization \cite{Ioffe2015} (all layers), rectified linear units (all layers), max pooling (first layer) and global average pooling (last layer). In the residual blocks that do not change the dimensionality of their inputs, identity shortcut connections are used. In the remaining residual blocks, convolutional shortcut connections are used. In contrast to \cite{He2015}, the number of convolutional kernels are halved.

Similar to \cite{Guclu2016}, the difference between the auditory stream and the visual stream is that inputs, convolutional/pooling kernels and strides of the auditory stream are one-dimensional whereas those of the visual stream are two-dimensional if the number of channels are ignored. That is:
\begin{itemize}
\item An $n ^ 2 \times 1 \times 1$ input of the auditory stream corresponds to an $n \times n \times m$ input of the visual stream.
\item An $n ^ 2 \times 1 \times m$/$n ^ 2 \times 1$ convolutional/pooling kernel of the auditory stream corresponds to an $n \times n \times m$/$n \times n$ convolutional/pooling kernel  of the visual stream.
\item An $n ^ 2 \times 1$ stride of the auditory stream corresponds to an $n \times n$ stride of the visual stream.
\end{itemize}
where $m$ is the number of channels.

Outputs of the auditory stream and the visual stream are merged in an audiovisual stream. The audiovisual stream comprises a fully-connected layer. The fully-connected layer is followed by hyperbolic tangent units. Outputs of the audiovisual stream are scaled to [0, 1].

\subsection{Training}

We used Adam~\cite{Kingma2014} with initial $\alpha = 0.0002$, $\beta_1 = 0.5$, $\beta_2 = 0.999$, $\epsilon = 10^{-8}$ and mini-batch size = 32 to train the network by iteratively minimizing the mean absolute error loss function between the target traits and the predicted traits for 900 epochs. We initialized the biases/weights as in \cite{He2014} and reduced $\alpha$ by a factor of 10 after every 300 epochs. Each training video clip was processed as follows:

\begin{itemize}
\item The audio data and the visual data of the video clip are extracted.
\item A random 50176 sample temporal crop of the audio data is fed into the auditory stream. The activities of the penultimate layer of the auditory stream are temporally pooled.
\item A random 224 pixels $\times$ 224 pixels spatial crop of a random frame of the visual data is randomly flipped in the left/right direction and fed into the visual stream. The activities of the penultimate layer of the visual stream are spatially pooled.
\item The pooled activities of the auditory stream and the visual stream are
concatenated and fed into the fully-connected layer.
\item The fully-connected layer outputs five continuous prediction values between the range [0, 1] corresponding to each trait for the video clip.
\end{itemize}

\subsection{Validation/Test}

Each validation/test video clip was processed as follows:

\begin{itemize}
\item The audio data and the visual data of the video clip are extracted.
\item The entire audio data are fed into the auditory stream. The activities of the penultimate layer of the auditory stream are temporally pooled (see below note).
\item The entire visual data are fed into the visual stream one frame at a time. The activities of the penultimate layer of the visual stream are spatiotemporally pooled (see below note).
\item The pooled activities of the auditory stream and the visual stream are
concatenated and fed into the fully-connected layer.
\item The fully-connected layer outputs five continuous prediction values between the range [0, 1] corresponding to each trait for the video clip.
\end{itemize}

It should be noted that the network can process video clips of arbitrary sizes since the penultimate layers of the auditory stream and the visual stream are followed by global average pooling.

\section{Results}
We evaluated the network on the dataset that was released as part of the ChaLearn First Impressions Challenge\footnote{\url{http://gesture.chalearn.org}.} \cite{Lopez2016}. The dataset consists of 10000 15-second-long video clips that were drawn from YouTube\footnote{\url{http://www.youtube.com/}.}, of which 6000 were used for training, 2000 were used for validation and 2000 were used for test. The video clips were annotated with the Big Five personality traits (i.e. openness to experience, conscientiousness, extraversion, agreeableness, and neuroticism) by Amazon Mechanical Turk\footnote{\url{http://www.mturk.com/}.} workers. Each trait was represented with a value between the range [0, 1].

The video clips were preprocessed by temporally resampling the audio data to 16000 Hz as well as spatiotemporally the video data to 456 pixels $\times$ 256 pixels and 25 frames per second.

\begin{figure}[!ht]
  \centering
    \includegraphics[width=1\textwidth]{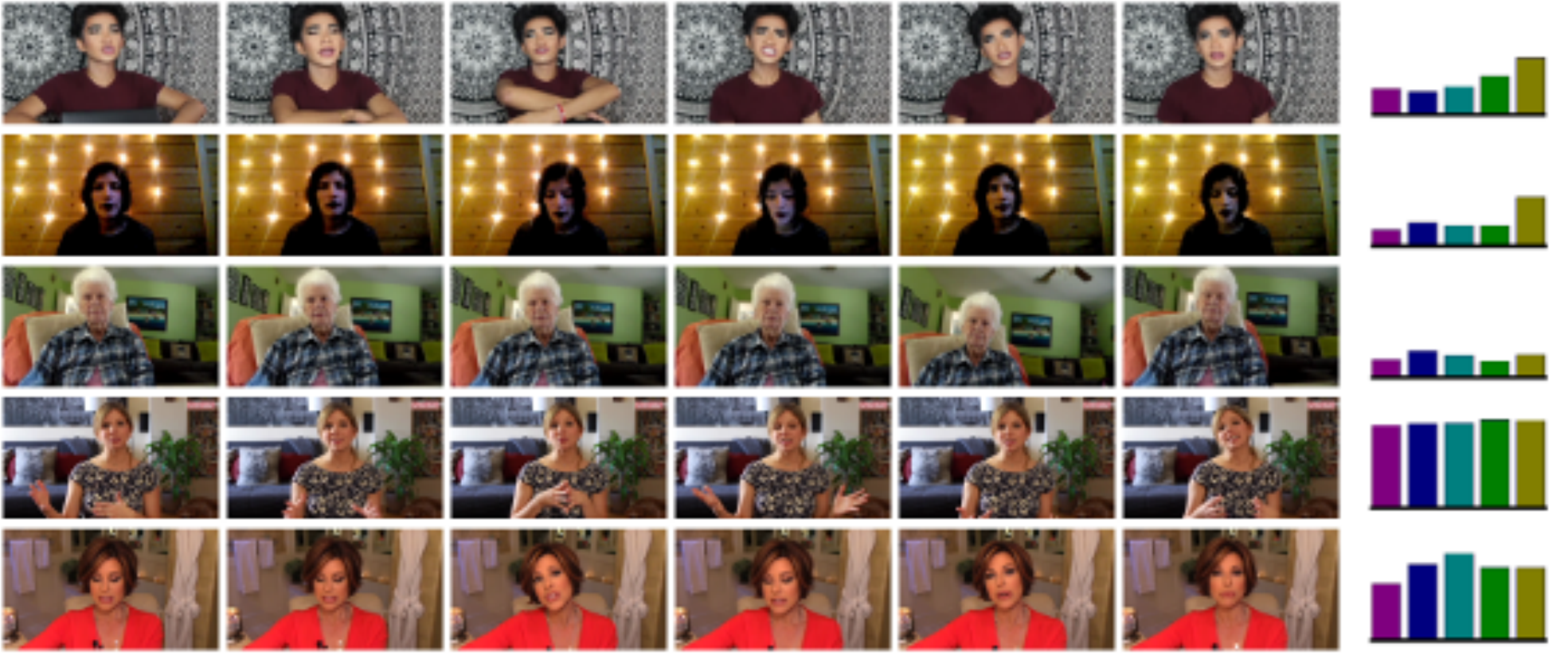}
  \caption{Example thumbnails of the videos of five people and the corresponding predicted personality traits. Each trait takes a value between [0, 1]. Each color represents a trait. From left to right: Openness, agreeableness, conscientiousness, neuroticism and extraversion.}
  \label{figure_2}
\end{figure}

We implemented the network in Chainer \cite{Tokui2015} with CUDA and cuDNN. Most of the processing took place on a single chip of an Nvidia Tesla K80 GPU accelerator \footnote{The implementation is available at \url{https://github.com/yagguc/deep_impression}.}. Processing took approximately 50 milliseconds per training example and 2.7 seconds per validation/test example on a single chip of an Nvidia Tesla K80 GPU accelerator. Figure \ref{figure_2} shows five validation examples and the corresponding predictions.

Accuracy was defined as 1 - mean absolute error. We report the validation accuracy of the network after 300, 600 and 900 epochs of training (Table \ref{table_1}). Average validation accuracy of the network increased as a function of number of epochs of training with the highest average validation accuracy of 0.9121. We report also the test accuracy of the network after 900 epochs of training, which won the third place in the challenge, and compare it with those of the models that won the first two places in the challenge (Table \ref{table_2}).

\begin{table}[]
\centering
\caption{Validation accuracies of the challenge model after 300, 600 and 900 epochs of training.}
\label{table_1}
\resizebox{\textwidth}{!}{%
\begin{tabular}{@{}ccccccc@{}}
\toprule
Epoch & \multicolumn{6}{c}{Validation accuracy} \\ \midrule
\multicolumn{1}{l}{} & Average & Openness & Agreeableness & Conscientiousness & Neuroticism & Extraversion \\ \midrule
300 & 0.906461 & 0.905451 & 0.911128 & 0.902121 & 0.907886 & 0.905721 \\
600 & 0.911929 & 0.911924 & \textbf{0.915610} & 0.911717 & \textbf{0.909891} & \textbf{0.910503} \\
900 & \textbf{0.912132} & \textbf{0.911983} & 0.915466 & \textbf{0.913077} & 0.909705 & 0.910429 \\ \bottomrule
\end{tabular}%
}
\end{table}

\begin{table}[]
\centering
\caption{Test accuracies of the models that won the first three places in the challenge.}
\label{table_2}
\resizebox{\textwidth}{!}{%
\begin{tabular}{@{}ccccccc@{}}
\toprule
Rank & \multicolumn{6}{c}{Test accuracy} \\ \midrule
\multicolumn{1}{l}{} & Average & Openness & Agreeableness & Conscientiousness & Neuroticism & Extraversion \\ \midrule
1 \cite{Zhang2016} & \textbf{0.912968037541} & \textbf{0.91237757} & \textbf{0.91257098} & \textbf{0.91663743} & \textbf{0.9099631} & 0.91329111 \\
2 \cite{Subramaniam2016} & 0.912062557634 & 0.91167725 & 0.91186694 & 0.91185413 & 0.90991703 & \textbf{0.91499745} \\
3 (ours) & 0.910932616931 & 0.91108539 & 0.91019226 & 0.91377735 & 0.90890031 & 0.91070778 \\
\multicolumn{7}{c}{\ldots} \\
10 & 0.875888740066 & 0.87026111 & 0.88423626 & 0.87270874 & 0.87526563 & 0.87697196 \\ \bottomrule
\end{tabular}%
}
\end{table}

\section{Post challenge models}

For completeness, we briefly report our preliminary work on two models that we have evaluated after the end of the challenge.

First, we separately fine-tuned the original DNN after 300 epochs of training for each trait. Everything about the fine-tuned DNNs (i.e. architecture, training and validation/test) were the same with the original DNN except for their fully-connected layers that output one value rather than five values.

Second, we trained a recurrent neural network (RNN) on top of the original network. The RNN comprised two layers of 512 long short-term memory units \cite{Hochreiter1997} and one layer of five linear units. At each time point, the RNN took as input the layer 5 features of a second-long video clip and the output of the RNN was the predicted traits. Dropout \cite{Srivastava2014} was used to regularize the hidden layers.

We used Adam to train the model by iteratively minimizing the mean absolute error loss function between the target traits and the predicted traits at each time point. Backpropagation was truncated after every 15 time points. Once the model was trained, the predicted traits were averaged over the entire video clip.

Table \ref{table_3} shows the validation accuracy of the post challenge models. While the post challenge models failed to outperform the challenge model to a large extent, we strongly believe that variants thereof have the potential to do so and will be the subject matter of future work.

\begin{table}[]
\centering
\caption{Validation accuracies of the challenge model and the post challenge models.}
\label{table_3}
\resizebox{\textwidth}{!}{%
\begin{tabular}{@{}ccccccc@{}}
\toprule
Model & \multicolumn{6}{c}{Validation accuracy} \\ \midrule
\multicolumn{1}{l}{} & Average & Openness & Agreeableness & Conscientiousness & Neuroticism & Extraversion \\ \midrule
DNN & 0.912132 & \textbf{0.911983} & 0.915466 & 0.913077 & \textbf{0.909705} & 0.910429 \\
5 $\times$ DNN & 0.911987 & 0.911522 & 0.915413 & 0.913211 & 0.909062 & 0.910727 \\
DNN + RNN & \textbf{0.912158} & 0.911676 & \textbf{0.915761} & \textbf{0.913300} & 0.909056 & \textbf{0.910996} \\ \bottomrule
\end{tabular}%
}
\end{table}

\section{Conclusion}

In this study, we presented our approach and results that won the third place in the ChaLearn First Impressions Challenge. Summarizing, we developed and trained an audiovisual deep residual network for predicting the apparent personality traits of people in an end-to-end manner. This approach enabled us to obtain very high performance for all traits while exploiting the similarities between the organization of the personality traits and the deep neural networks in terms of the hierarchical organization and circumventing extensive analyses for identifying/designing relevant features for the task of apparent personality traits prediction. Our results demonstrate the potential of deep neural networks in the field of automatic (perceived) personality prediction. Future work will focus on the extensions of the current work with recurrent neural networks and language models as well as identifying the factors that drive first impressions.

\clearpage

\bibliographystyle{splncs03}
\bibliography{main}
\end{document}